\documentclass[conference]{IEEEtran}

\IEEEoverridecommandlockouts

\usepackage{cite}
\usepackage{amsmath,amssymb,amsfonts}
\usepackage{threeparttable}

\newcommand{\linebreakand}{%
  \end{@IEEEauthorhalign}
  \hfill\mbox{}\par
  \mbox{}\hfill\begin{@IEEEauthorhalign}
}
\makeatother 

\usepackage{graphicx}
\usepackage{textcomp}
\usepackage[dvipsnames]{xcolor}
\usepackage{subfigure}
\usepackage{subcaption}
\usepackage{enumerate}
\usepackage{algorithm}
\usepackage{float}
\usepackage[noend]{algpseudocode}
\usepackage{diagbox}
\usepackage{booktabs}
\usepackage{adjustbox}
\usepackage{comment}
\usepackage{multirow}
\usepackage{pifont}
\usepackage{hyperref}
\usepackage{makecell}
\usepackage{tabularx}
\usepackage{makecell}

\DeclareRobustCommand{\[}{\begin{equation}}
\DeclareRobustCommand{\]}{\end{equation}}

\def\BibTeX{{\rm B\kern-.05em{\sc i\kern-.025em b}\kern-.08em
    T\kern-.1667em\lower.7ex\hbox{E}\kern-.125emX}}

\usepackage[final,commandnameprefix=always]{changes}
\definechangesauthor[color=BrickRed,name=Tozammel]{TH}

\begin{document}


\title{Modeling and Measuring Redundancy in Multisource Multimodal Data for Autonomous Driving}

\author{\IEEEauthorblockN{Yuhan Zhou}
\IEEEauthorblockA{\textit{Dept. of Information Science} \\
\textit{University of North Texas}\\
Denton, Texas, USA \\
yuhanzhou@my.unt.edu}

\and
\IEEEauthorblockN{Mehri Sattari}
\IEEEauthorblockA{\textit{Dept. of Information Science} \\
\textit{University of North Texas}\\
Denton, Texas, USA \\
MehriSattari@my.unt.edu}

\and
\IEEEauthorblockN{Haihua Chen}
\IEEEauthorblockA{\textit{Dept. of Data Science} \\
\textit{University of North Texas}\\
Denton, Texas, USA \\
Haihua.Chen@unt.edu}

\and
\IEEEauthorblockN{Kewei Sha}
\IEEEauthorblockA{\textit{Dept. of Data Science} \\
\textit{University of North Texas}\\
Denton, Texas, USA \\
Kewei.Sha@unt.edu}
}

\maketitle

\thispagestyle{plain}
\pagestyle{plain}

\begin{abstract} 
Next-generation autonomous vehicles (AVs) rely on large volumes of multisource and multimodal ($M^2$) data to support real-time decision-making. In practice, data quality (DQ) varies across sources and modalities due to environmental conditions and sensor limitations, yet AV research has largely prioritized algorithm design over DQ analysis. This work focuses on redundancy as a fundamental but underexplored DQ issue in AV datasets. Using the nuScenes and Argoverse 2 (AV2) datasets, we model and measure redundancy in multisource camera data and multimodal image–LiDAR data, and evaluate how removing redundant labels affects the YOLOv8 object detection task. Experimental results show that selectively removing redundant multisource image object labels from cameras with shared fields of view improves detection. In nuScenes, mAP${50}$ gains from $0.66$ to $0.70$, $0.64$ to $0.67$, and from $0.53$ to $0.55$, on three representative overlap regions, while detection on other overlapping camera pairs remains at the baseline even under stronger pruning. In AV2, $4.1$–$8.6\%$ of labels are removed, and mAP${50}$ stays near the $0.64$ baseline. Multimodal analysis also reveals substantial redundancy between image and LiDAR data. These findings demonstrate that redundancy is a measurable and actionable DQ factor with direct implications for AV performance. This work highlights the role of redundancy as a data quality factor in AV perception and motivates a data-centric perspective for evaluating and improving AV datasets. Code, data, and implementation details are publicly available at: \url{https://github.com/yhZHOU515/RedundancyAD}. 
\end{abstract}

\begin{IEEEkeywords}
Autonomous vehicles, data-centric AI, data quality, multimodal, multisource
\end{IEEEkeywords}

\section{Introduction}

Autonomous vehicles (AVs) influence human daily life in transportation mobility and driving safety \cite{guo2024advances, liuautonomous}. It also pushes the frontier of data engineering in data fusion, data simulation, and real-time data processing. The real-world challenge in AVs lies in modeling the highly dynamic environment, which contains long-tail risky cases and uncertainty \cite{han2023collaborative, Chen2024E2E, guo2024advances, li2023open, Liang_2024_AIDE}. Current AV frameworks and data engineering approaches mainly emphasize benchmark development and evaluation, and model-centric architectural designs \cite{kim2025automatic}. However, raw compute or larger neural networks cannot eliminate risk when the input data from various sensors is noisy, inconsistent, or biased. Despite the significant role of data quality (DQ) in assuring the performance, efficiency, and reliability of AVs, these strategies not only overlook DQ evaluation but also treat all AV tasks with similar DQ requirements \cite{li2024datacentricAV}. Consequently, the lack of DQ evaluation and task-centric awareness prevents AV systems from feeding quality diagnostics back into the loop and iterating on performance \cite{sun2020scalability, Chen2024E2E, li2023open}.

However, the data quality work in AV is still restrained due to the absence of a systematic evaluation of data quality dimensions under specific tasks \cite{li2022traffic}, the general standards and definitions for ``high-quality” data \cite{li2023open}, and limited evidence and explanation on data quality impacts \cite{goknil2023systematic}. Prior data work in AV focused more on dataset characteristics \cite {guo2024advances, li2024datacentricAV, li2023open, liu2024DatasetSurvey}, such as weather conditions, geospatial coverage, sensor types, dataset sizes, and instance diversity \cite{sun2020scalability, caesar2020nuscenes, geiger2013vision, alibeigi2023zenseact}. In the meantime, few papers have discussed data quality evaluation specifically for AV pipelines \cite{zhou2025novel}. In real-world settings, the quality of different data sources and modalities usually varies due to unexpected environmental factors or sensor issues \cite{zhang2024multimodal}. Therefore, it is crucial to research the dynamically varying quality of multisource and multimodal ($M^2$) data and its impact on model performance.

\begin{figure*}[h]
    \centering
    \includegraphics[width=1\linewidth]{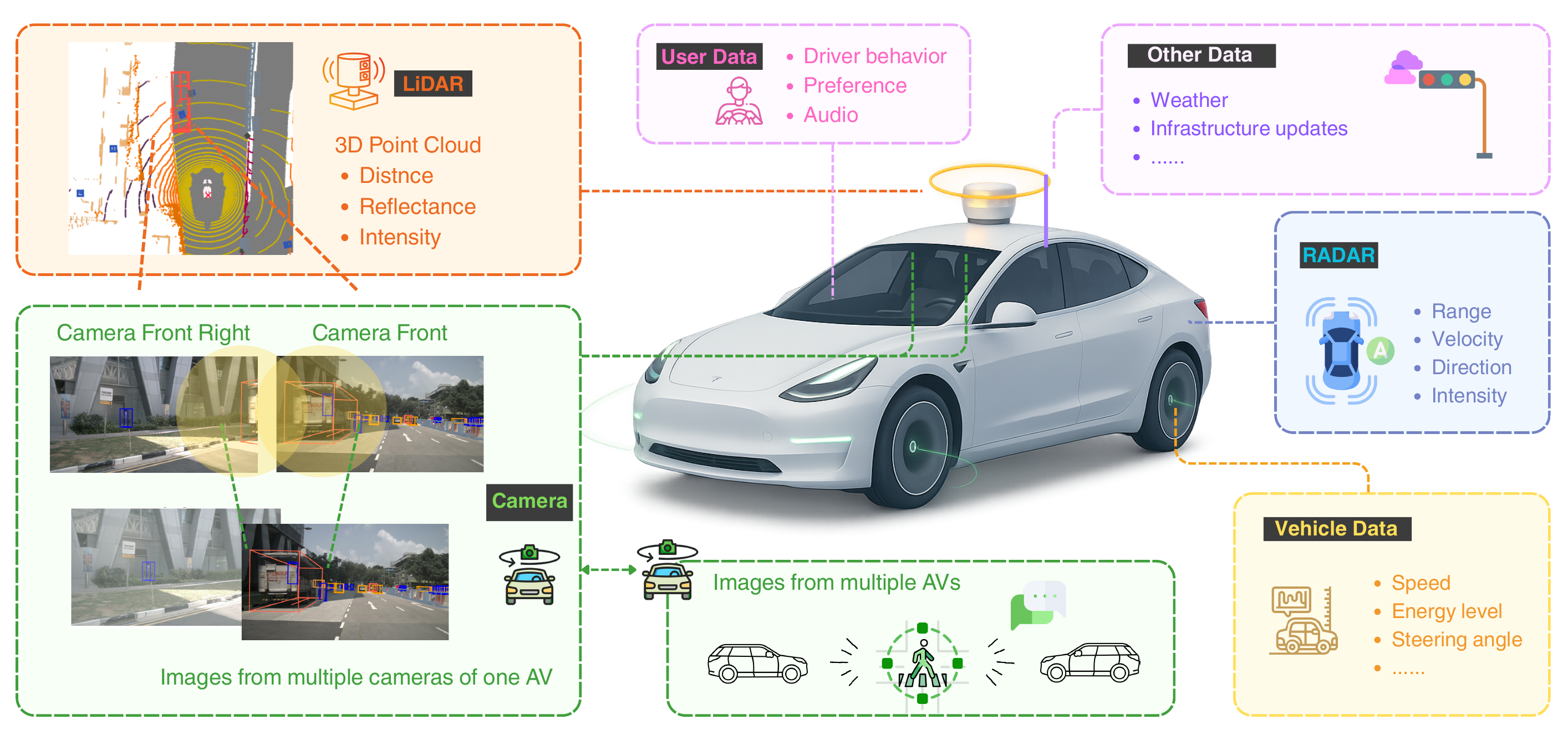}
    \caption{Illustration of multisource and multimodal data in autonomous vehicles. LiDAR point clouds, multi-camera images, RADAR, vehicle dynamics, user-related data, and external contextual inputs feed into the AV system. Redundancy arises from overlapping fields of view among multiple onboard cameras (green), as well as from multimodal sensing in which cameras and LiDAR (orange) observe the same objects. The rest colored blocks represent additional data sources and modalities.} 
    \label{fig:multi-source}
\end{figure*}

Fig. \ref{fig:multi-source} illustrates the $M^2$ data and the redundancy of an example autonomous vehicle. \textbf{Multisource} data are collected from heterogeneous onboard sensors (e.g., cameras, LiDAR, and RADAR), which offer complementary yet partially redundant representations of the same driving scene. These data are further integrated with external inputs such as GPS, traffic, infrastructure, and weather data, as well as data shared by connected autonomous vehicles (CAVs). The redundancy across multisource data improves situational awareness and system robustness by mitigating sensor uncertainty and failures, thereby supporting more reliable perception and decision making \cite{wang2023vimi}. \textbf{Multimodal} data combines signals captured through different sensing modalities. For example, images from cameras and point clouds from LiDAR can contribute complementary information about the same scene, enhancing the accuracy of AV tasks such as object detection and localization. 

Redundancy arises from multiple sensors on the same vehicle in autonomous driving (AD) \cite{sun2020advanced}. Although redundancy can improve reliability, its absence of explicit quantification poses major risks: duplicate detections increase computational costs and harm real-time efficiency \cite{Zhao_2023_ada3d, li2023reducing, Sun2023Task-Driven, xu2024dm3d}. Inconsistent redundant predictions can cause noise, degrading localization and confidence. Thus, modeling and measuring redundancy in a specific AV task is crucial, which can help spatial verification and fault tolerance without overwhelming the detection pipeline \cite{liu2021Multi-sensorRedundancy, qian20223d}. This motivates us to model and measure $M^2$ redundancy in the object detection (OD) task in AD. Based on the above, the research questions are formulated as follows.

\begin{itemize}
    \item \textbf{RQ1:} How to define and model redundancy in multisource and multimodal AV data for object detection?
    \item \textbf{RQ2:} How to measure redundancy and select the optimal data subset for object detection in AV?
    \item \textbf{RQ3:} How does redundancy removal affect object detection model performance?
\end{itemize}

\begin{figure*}[h]
    \centering
    \includegraphics[width=\linewidth]{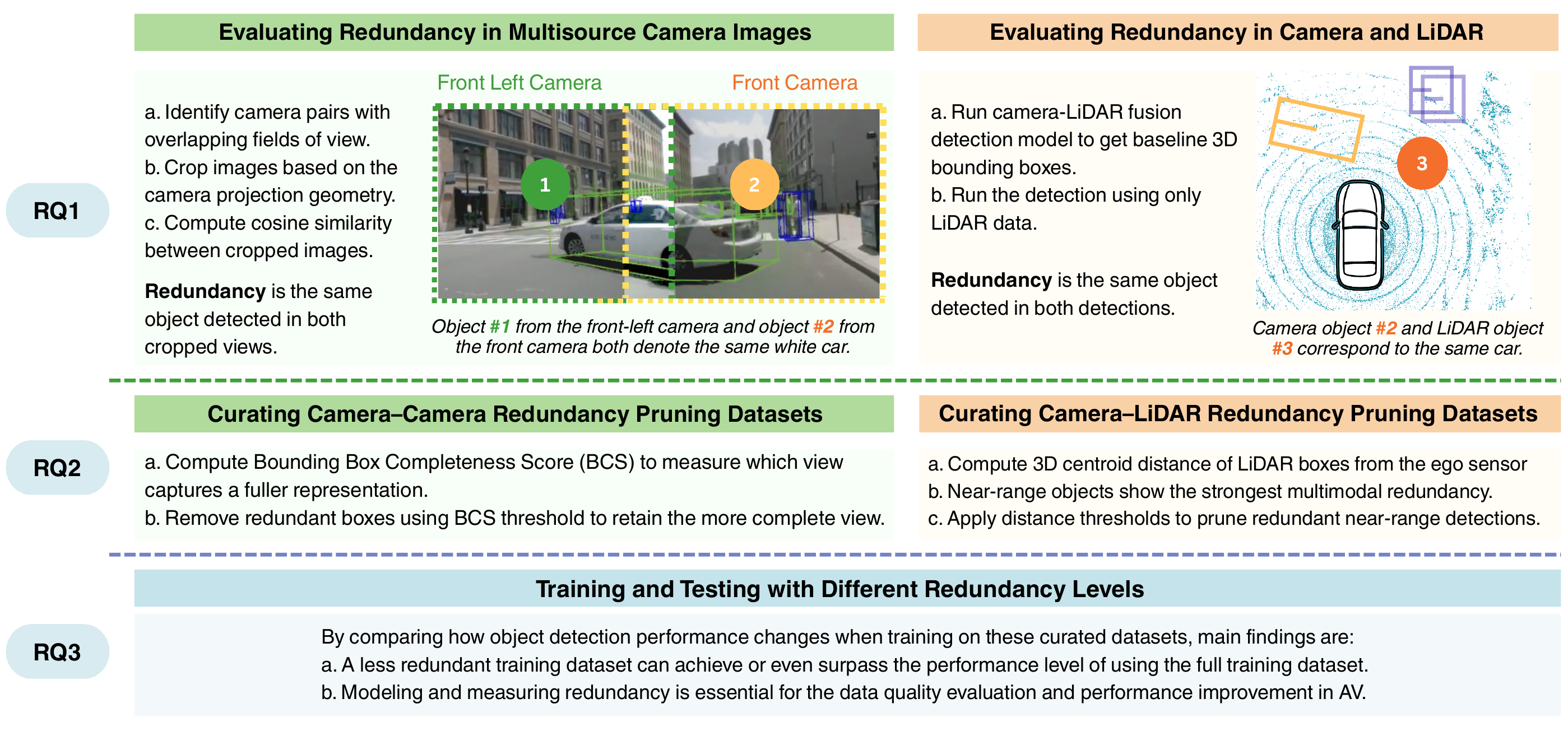}
    \caption{Illustration of research design, using nuScenes dataset as examples. We first quantify redundancy in multisource and multimodal autonomous driving datasets (RQ1), then design data pruning to remove redundant observations (RQ2), and finally evaluate how redundancy removal affects object detection performance (RQ3).}
    \label{fig: RD}
\end{figure*}

Fig. \ref{fig: RD} illustrates our overall research design, where each component directly addresses one of the three research questions. To answer RQ1, we evaluate redundancy across multisource (camera–camera) and multimodal (camera–LiDAR) observations in the object detection task. To answer RQ2, we investigate which redundant observations to remove and propose pruning strategies that retain the most complete object representations. To answer RQ3, we train and evaluate YOLOv8 using datasets curated with different redundancy levels, and analyze how redundancy removal influences detection performance and data efficiency.

The contributions of this paper can be summarized as follows. 
\begin{enumerate}
    \item For the first time, we model and measure redundancy in multisource and multimodal data in the object detection task. We design practical redundancy measurement strategies for camera–camera and camera–LiDAR data, highlighting the importance of data-quality research in autonomous driving.
    \item We propose a task-driven data selection method based on bounding box completeness and spatial overlap constraints. We validate that our redundancy measurement and selection pipeline is not dataset-specific but generalizable to AV benchmarks.
    \item We evaluate redundancy reduction on YOLOv8 object detection using nuScenes and Argoverse 2 (AV2). The results show that reducing redundancy can maintain or even improve detection performance.
    \item We reveal and analyze cross-modal redundancy between camera images and LiDAR point clouds, providing empirical evidence that multimodal redundancy is substantial and must be explicitly measured to balance robustness and efficiency.
\end{enumerate}

The rest of the paper is organized as follows. Section \ref{sec:related-work} provides a literature review on the $M^2$ data and data quality issues in AV, and specifically, the redundancy evaluation. Section \ref{sec:method} introduces the research design, object detection task, the model, and performance metrics. Section \ref{sec: exp} provides details of the datasets and experiments. Section \ref{sec: result} and Section \ref{sec: discussion} present the experiment results and discussions, respectively. Finally, Section \ref{sec:con} concludes our work and puts forward future directions.

\section{Related Work}
\label{sec:related-work}

\subsection{Multisource and Multimodal data in Autonomous Driving}

At the \textbf{multisource} level, AVs gather information from various sensors, such as LiDAR, cameras, and RADAR, as well as other feeds including user behavior data, weather reports, and infrastructure updates. Since one AV can be equipped with more than one camera, the front, back, and side cameras are considered different data sources \cite{zhou2020end, munsif2025multi}. The integration of multisource data can validate data and reduce the impact of errors or noises from individual sensors, which enables a more comprehensive view of the driving environment and a more accurate and robust performance \cite{zhang2022sensor, song2019apollocar3d, wang2023vimi}, enhancing the system’s situational awareness, reliability, and robustness.

\textbf{Multimodal} data in AV refers to information obtained through different modes of data collection, mainly including various types of sensory data such as visual (camera), auditory (user), and spatial (LiDAR, RADAR) data, as shown in Fig. \ref{fig:multi-source}. Each mode provides unique and complementary insights that, when combined, offer a more holistic understanding of the environment. Multimodal data fusion enables these systems to perform complex tasks such as object detection, localization, and navigation with higher accuracy and reliability by leveraging the strengths of each data modality.

Combining $M^2$ data can enhance performance and reliability in AV compared to a single data source or modality \cite{su2024adverse, hou2023fault}. Zhang et al. improved vehicle localization by integrating vehicle headlights and taillights detected from multiple cameras \cite{zhang2022night}. Sanchez et al. found that single-source LiDAR detectors generalize poorly, and multisource LiDAR training can improve domain generalization by over 10\% and accuracy by 5.3\% over any individual dataset \cite{sanchez2025cola}. When LiDAR points are sparse for tiny objects, a camera could help recover them by visual cues \cite{fu2024eliminating, wang2023scenes, wang2023od, yang2024deepinteraction++}.


\subsection{Data Quality in Autonomous Driving}

Data quality affects machine learning performance \cite{chen2021data}. High-quality training data enhances accuracy, while low-quality data can lead to biased or erroneous outcomes \cite{sambasivan2021everyone, garbage2020}. Recognizing this, scholars have developed frameworks \cite{rangineni2023analysis, Priestley2023, schwabe2024metric, rahal2025enhancing} and DQ dimensions, such as accuracy \cite{chen4979696enhancing}, completeness \cite{emmanuel2021survey}, consistency \cite{Li2021labelQuality, kong2025multi}, redundancy \cite{chen2023survey} and fairness \cite{Guha2024fair}. 


Data Quality in autonomous driving is the backbone for assuring performance, efficiency, reliability, and robustness. Particularly, $M^2$ data in CAVs are complex to handle, making the DQ evaluation more significant and challenging. In real-world settings, the quality of different modalities and data sources usually varies due to unexpected environmental factors or sensor issues \cite{zhang2024multimodal}. This brings major challenges (but not limited to): (1) mitigate the underlying influence of arbitrary noise, bias, and discrepancy, (2) enhance data quality in terms of different dimensions, (3) adapt the quality of dynamically varying nature of $M^2$ data. 

However, there is a lack of systematic evaluation of data quality dimensions under specific tasks \cite{li2022traffic}. Most existing studies primarily focus on dataset characteristics \cite{liu2024DatasetSurvey, li2024datacentricAV}, such as geospatial coverage, sensor types, and instance diversity \cite{li2022traffic}. Few studies have defined specific DQ dimensions of $M^2$ data in AVs from a task-centric perspective and answered the question of how to select useful data.



\subsection{Redundancy in Autonomous Driving Datasets}

By having multiple sources of similar information, the system can validate data, reducing the impact of errors or noise from individual sensors. When one sensor fails or generates deviated data, the other sensors can still perform \cite{liu2021Multi-sensorRedundancy, qian20223d, he2020sensorRedun}. This enables a more accurate and robust performance \cite{zhang2022sensor, song2019apollocar3d}. In object detection, multiple sensors are set up to obtain comprehensive and reliable environmental information \cite{wang2023od, wang2023scenes}. 

Redundancy can negatively impact the AD system. Firstly, it will decrease the efficiency in computational time and storage \cite{Zhao_2023_ada3d, li2023reducing, Sun2023Task-Driven, xu2024dm3d}. For example, when the weather conditions are good and clear, there is less need to keep high-overlapping information from LiDAR and cameras. The models trained on such data would have little information gain and performance improvement \cite{duong2022active, ju2022extending}. The redundant information includes the temporal correlation between point cloud scans, similar urban environments, and the symmetries in the driving environment caused by driving in opposite directions at the same location \cite{duong2022active, li2023reducing}. As an example, after randomly removing 30\% of the redundant background cloud points, Zhao et al. found only a subtle drop in performance, which proves the necessity and efficiency of researching the redundancy data quality issues in the datasets \cite{Zhao_2023_ada3d}.

Evaluating cross-modal redundancy poses challenges. Each modality has a different informative level and contributes to different tasks \cite{lu2024crossprune}. Reducing redundancy is not simply removing duplicate records, as in previous single-modality data analysis tasks. It requires investigations on the sensor characteristics, task specificity, performance experiments, trade-off evaluation, and so on. Li et al. view redundancy as a content data quality issue and divide redundancy into element redundancy and data redundancy \cite{li2023quality}, quantifying how much one or several elements co-occur and repeat the same behavior for longer than the threshold. Duong et al. define redundancy as the similarity between pairs of point clouds resulting from geometric transformations of ego-vehicle movement and environmental changes \cite{duong2022active}. Rao et al. also calculated the similarity between the images and text pairs \cite{rao2020quality}. From an information science perspective, calculating mutual information is also an approach \cite{hadizadeh2024mutual}.

\section{Methodology}
\label{sec:method}

\subsection{Research Design}

\begin{figure*}[h]
    \centering
    \includegraphics[width=1\linewidth]{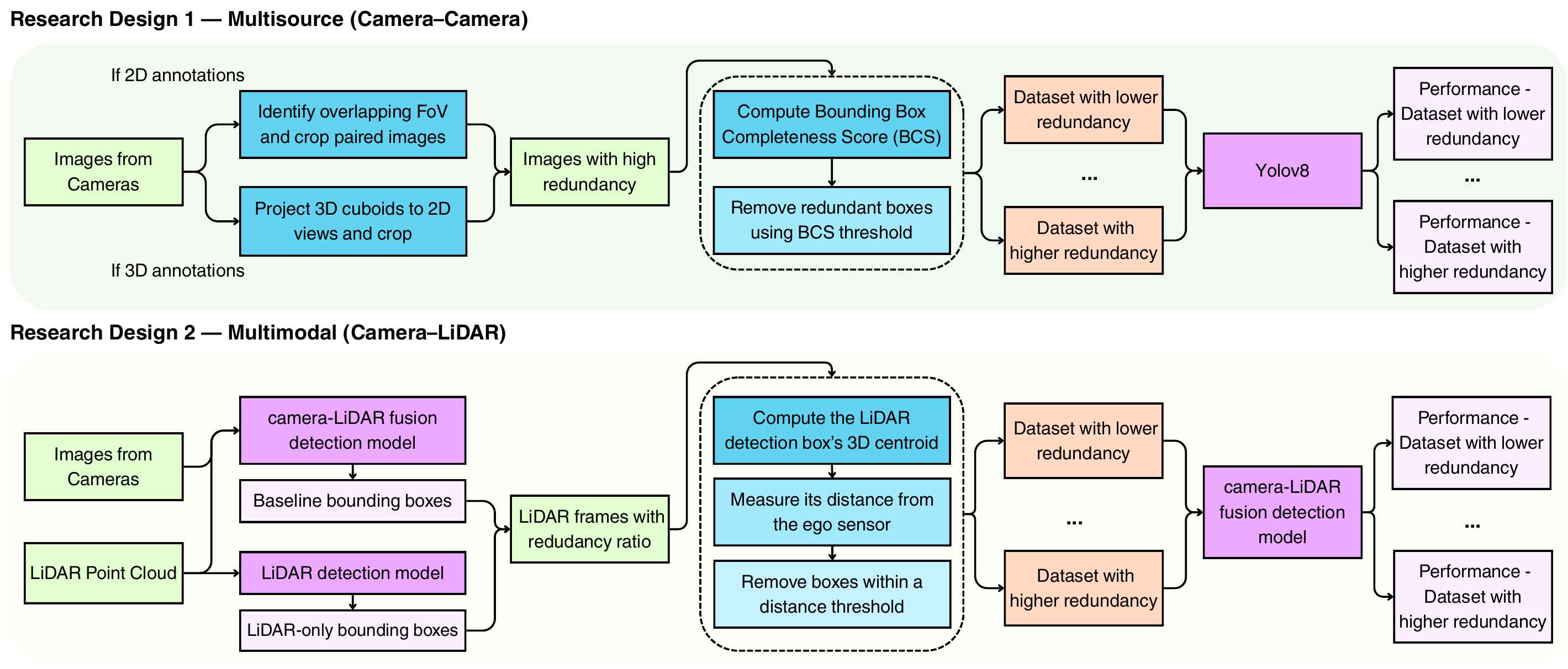}
    \caption{The workflow of the research design on redundancy evaluation. nuScenes emphasizes 2D labeling and AV2 adopts 3D annotations. Beyond dataset-specific configurations, this workflow can be generalized to other AD datasets.}
    \label{fig: workflow}
\end{figure*}

This section introduces the research design on redundancy evaluation across the multisource camera views and the multimodal camera and LiDAR data for the object detection task. Fig. \ref{fig: workflow} introduces the workflow of experiments. (i) For multisource redundancy evaluation, when 2D labels are present, we directly estimate overlapping fields of view (FoVs) between camera pairs and crop shared regions. For datasets that provide 3D annotations, we project 3D cuboids and identify object views based on projection overlap. We then design a Bounding-Box Completeness Score (BCS) to measure how fully each object instance is captured per view, and remove redundant camera observations by BCS thresholding. (ii) To measure multimodal redundancy, we compute the 3D centroid of the predicted box using LiDAR data and measure its ego-centric distance from the vehicle sensor origin. Rather than pruning by confidence alone, we remove LiDAR boxes within a foreground distance threshold where visual coverage is already strong, preserving observations for distant or visually incomplete objects.

The research design is guided by three considerations, consistent with our objective to model and measure redundancy in AD.
(i) Overlapping FoVs in multisource images produce duplicate 2D labels for the same object, requiring explicit spatial redundancy modeling.
(ii) In camera–LiDAR fusion, object importance is strongly conditioned on sensor range, motivating distance-aware redundancy measurement and pruning beyond IoU-based criteria.
(iii) Redundancy removal must be tailored to the specific task. To preserve the most informative instance per object in object detection, we adopt BCS to retain the most complete 2D camera observation in overlapping views and apply LiDAR box centroid-to-ego distance thresholds to filter temporally correlated 3D duplicates.

\subsection{Task Formulation}

Our task objective is to evaluate whether removing redundant observations of the same object across sources and modalities can improve object detection performance. The task is defined as learning a mapping
\begin{equation}
f_{\theta}: \mathcal{X} \rightarrow \mathcal{Y},
\end{equation}
where $\mathcal{X}$ denotes multisource or multimodal inputs (overlapping view pairs) and $\mathcal{Y}$ denotes object detection outputs.

\subsection{Redundancy Modeling and Evaluation}


\subsubsection{Modeling Redundancy in Multisource Data}

We consider overlapping camera views that can produce multiple observations of the same physical object. For each object instance $o_i$ at time step $t$, let $\mathcal{P}_t$ denote the set of overlapping camera-view pairs at time $t$ (obtained either from dataset-defined overlap pairs or from calibration-driven FoV overlap estimation). For a selected pair $(c_a, c_b)\in \mathcal{P}_t$, the input is a pair of image regions extracted from the overlapping FoVs of the two cameras,
\begin{equation}
\mathcal{X}^{\text{src}}_{t,i} =
\big( \tilde{I}^{(a)}_{t,i},\; \tilde{I}^{(b)}_{t,i} \big),
\end{equation}
where $\tilde{I}^{(a)}_{t,i}$ and $\tilde{I}^{(b)}_{t,i}$ are RGB image regions from cameras $c_a$ and $c_b$ corresponding to their overlapping FoV and potentially observing the same physical object.

Given an input $\mathcal{X}_{t,i}$, the detector predicts
\begin{equation}
\hat{\mathcal{Y}}_{t,i} = \{ (\hat{b}_i, \hat{c}_i, \hat{s}_i) \},
\end{equation}
where $\hat{b}_i$ denotes the bounding box, $\hat{c}_i$ the class label, and $\hat{s}_i$ the confidence score. 

\subsubsection{Modeling Redundancy in Multimodal Data}

The input consists of paired frames from image and LiDAR modalities,
\begin{equation}
\mathcal{X}^{\text{mod}}_{t,i} =
\big( I_{t,i},\; L_{t,i} \big),
\end{equation}
where $I_{t,i}$ and $L_{t,i}$ denote synchronized image and LiDAR observations. The output is the same as the multisource data redundancy evaluation.

\subsection{Model for Object Detection}

Our redundancy evaluation is conducted under the object detection task, where duplicated inputs can directly influence localization quality, model bias, and training efficiency. OD aims to jointly localize and classify objects of interest, enabling autonomous driving systems to parse dynamic road scenes and support safety-critical decision-making \cite{wang2023od}. OD pipelines typically rely on heterogeneous sensors, including cameras, LiDAR, and RADAR. Given $M^2$ input, the model outputs 2D and/or 3D bounding boxes for detected objects. Representative OD architectures include CNN-based detectors (e.g., YOLO \cite{terven2023yolo}, VoxelNet \cite{zhou2018voxelnet}, PointNet \cite{qi2017pointnet}), transformer models \cite{li2022ViT}, and multimodal fusion methods such as BEVFusion \cite{liu2023bevfusion} and BEV-space camera–LiDAR frameworks.

To evaluate the impact of redundancy pruning, we require a model with a strong and stable baseline accuracy to observe performance shifts from redundant supervision. We therefore adopt YOLOv8, a high-performance, real-time CNN detector on AD benchmarks. YOLOv8 was trained with a batch size of 16 and 50 epochs for the object detection task for evaluation. 

\subsection{Performance Evaluation Metrics}

Object detection performance is commonly measured using mean Average Precision (mAP) and recall. In this work, mAP$50$ is used to assess bounding box alignment with ground-truth labels at $\geq50\%$ Intersection over Union (IoU), while recall measures the fraction of true objects successfully detected. 

Given multiple redundancy levels $\{ r_1, r_2, \ldots, r_m \}$ constructed under both multisource and multimodal settings, we compare
\begin{equation}
\text{Performance}(\mathcal{X}^{(r_1)}) \quad \text{vs.} \quad \text{Performance}(\mathcal{X}^{(r_2)}),
\end{equation}
across these settings to evaluate the effect of redundancy on object detection performance.

\section{Experiments}
\label{sec: exp}

This section introduces the datasets and experiments on redundancy evaluation to validate the proposed methodology. 

\subsection{Dataset}

The datasets used are nuScenes-mini, nuScenes-in-KITTI, and Argoverse 2. We select nuScenes and Argoverse 2 due to their suitability for this redundancy evaluation study. First of all, they are well-known, large-scale, multisensor AD benchmarks. Secondly, both datasets include public calibration files and dense annotations, which enable a robust evaluation of redundancy across spatial and temporal axes. nuScenes provides a 360-degree multi-camera rig with substantial inter-view FoV overlap and densely annotated 3D boxes, making it well-suited for studying camera–camera redundancy and cross-modal pruning in fusion models. The details are introduced in Table \ref{tab:datasets} and as follows. 

\begin{table*}[htbp]
\centering
\caption{Summary of datasets used for redundancy evaluation}
\label{tab:datasets}
\begin{tabular}{p{2.0cm} p{0.7cm} p{3.7cm} p{4.0cm} p{2.5cm} p{2.2cm}}
\toprule
\textbf{Dataset} & \textbf{Year} & \textbf{Sensors} & \textbf{Annotations} & \textbf{Size} & \textbf{Modalities} \\
\midrule

nuScenes-mini \cite{caesar2020nuscenes} &
2020 &
6 cameras (12\,Hz), 1 LiDAR, RADAR, GPS/IMU &
3D bounding boxes, object attributes, tracking, HD maps &
10 scenes, 404 frames &
Multisource camera images \\

\midrule

nuScenes-in-KITTI \cite{NuScenes-in-Kitti}&
2021 &
Cameras, 1 LiDAR, GPS/IMU &
2D/3D bounding boxes, calibration, tracking &
85 scenes in KITTI format &
Image and LiDAR \\

\midrule

Argoverse~2 \cite{wilson2023argoverse}&
2023 &
9 cameras, 2 LiDARs (10\,Hz), GPS/IMU &
3D bounding boxes, tracking, map annotations, motion labels &
1,000 scenes, 20,000 LiDAR sequences &
Multisource camera images \\

\bottomrule
\end{tabular}
\end{table*}

\paragraph{nuScenes-mini}
The nuScenes dataset is a large-scale, multimodal dataset designed for autonomous driving research \cite{caesar2020nuscenes}. It includes 360-degree sensor coverage with data collected from six cameras, five RADARs, and one LiDAR sensor, along with high-definition maps and detailed object annotations. The cameras are of 12Hz capture frequency. The full dataset consists of 1,000 driving scenes, providing a rich resource for perception tasks such as object detection, tracking, and scene understanding. The nuScenes-mini dataset is a smaller subset containing 10 scenes and 404 frames with the same sensor setting and annotation structure. 

\paragraph{nuScenes-in-KITTI}
nuScenes-in-KITTI \cite{NuScenes-in-Kitti} is a nuScenes version that is compatible with the KITTI format \cite{Geiger2012kitti}. KITTI is a collection of synchronized sensor recordings from a car equipped with cameras, LiDAR, and GPS/IMU, driving through city, residential, campus, rural, and highway scenes in Germany.

\paragraph{Argoverse 2}
Argoverse 2 is designed to support robust 3D perception, tracking, and motion forecasting research \cite{wilson2023argoverse}. It provides multimodal sensor data collected across diverse urban environments. It consists of 1,000 annotated driving sequences with dense 3D bounding box annotations and tracking labels. The dataset’s scale, sensor redundancy, and diverse scenarios make it well-suited for evaluating redundancy in both image and LiDAR modalities. 

\subsection{Evaluating Redundancy in Multisource data in nuScenes}

This experiment focuses on the multisource camera images. As shown in Fig. \ref{fig: camera}, the nuScenes camera setup has six pairs of overlapping FoV. These overlapping parts indicate areas where cross-camera redundancy may occur. In other words, the front right camera and the back left camera can not capture the same instance simultaneously. Therefore, these six pairs are our research focus. To illustrate our framework by redundancy in this kind of multisource data, we experiment in the following steps:

\begin{figure}
    \centering
    \includegraphics[width=0.75\linewidth]{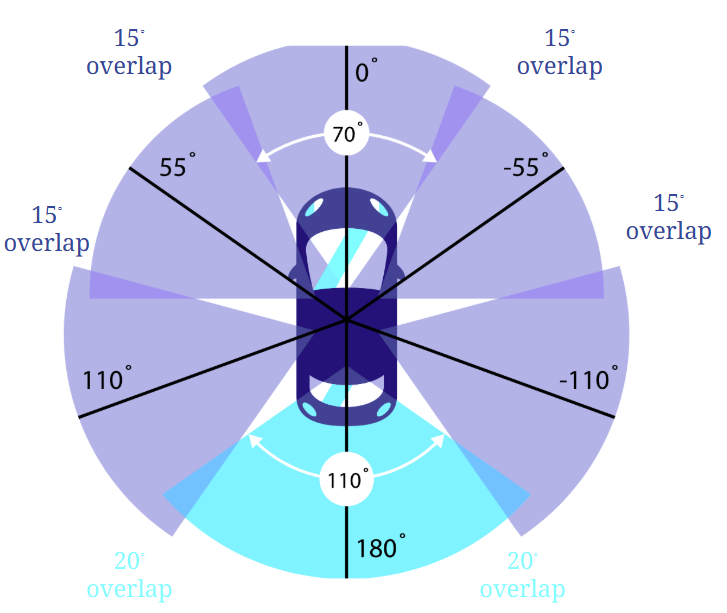}
    \caption{Camera settings and overlapping FoVs in nuScenes. For datasets with different sensor layouts, the same redundancy modeling and measurement approach remains applicable by adjusting the geometric parameters.}
    \label{fig: camera}
\end{figure}

(a) Identify the overlapping FoV based on the nuScenes dataset sensor setting: 

\begin{itemize}
    \item \textbf{Pair 1}: \texttt{CAM\_FRONT} and \texttt{CAM\_FRONT\_RIGHT} overlap by 15° 
    \item \textbf{Pair 2}: \texttt{CAM\_FRONT} and \texttt{CAM\_FRONT\_LEFT} overlap by 15° 
    \item \textbf{Pair 3}: \texttt{CAM\_FRONT\_RIGHT} and \texttt{CAM\_BACK\_RIGHT} overlap by 15°
    \item \textbf{Pair 4}: \texttt{CAM\_FRONT\_LEFT} and \texttt{CAM\_BACK\_LEFT} overlap by 15°
    \item \textbf{Pair 5}: The rear camera \texttt{CAM\_BACK} overlaps with \texttt{CAM\_BACK\_RIGHT} by 20°
    \item \textbf{Pair 6}: The rear camera \texttt{CAM\_BACK} overlaps with \texttt{CAM\_BACK\_LEFT} by 20°
\end{itemize}

\begin{figure}
    \centering
    \includegraphics[width=1.0\linewidth]{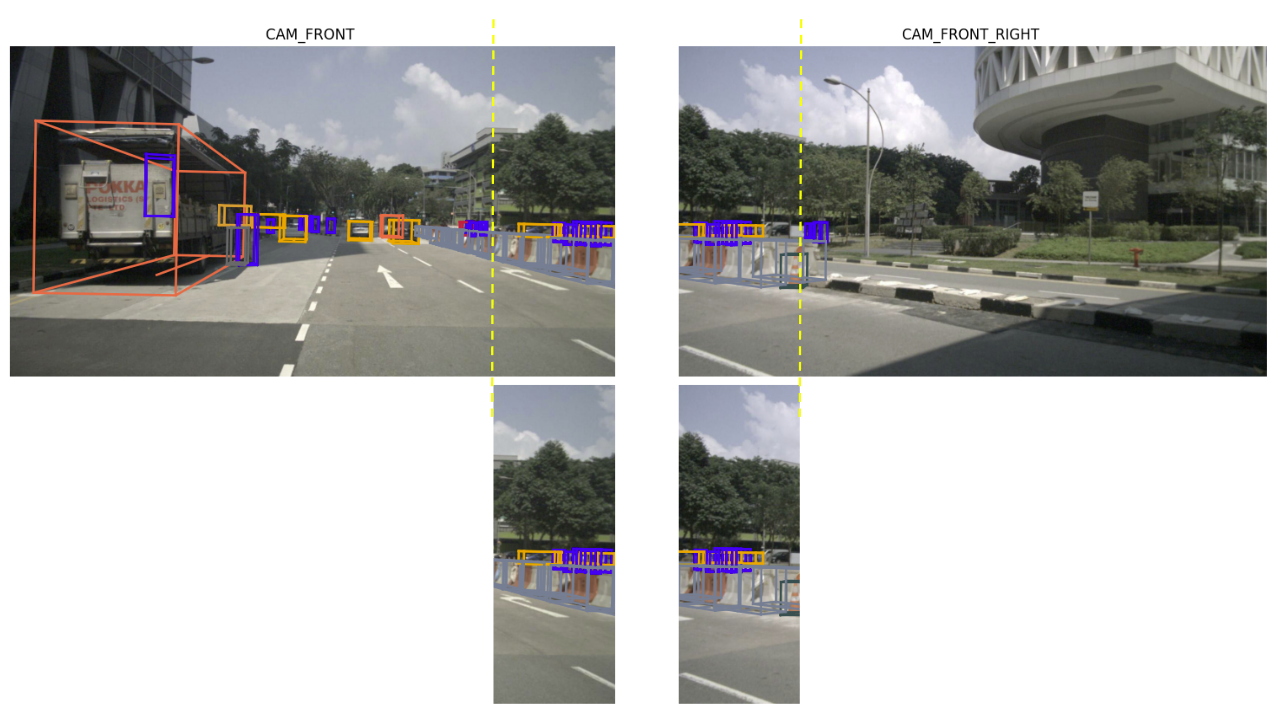}
    \caption{Illustration of cropping based on overlapping angles of the cameras.}
    \label{fig: crop}
\end{figure}

(b) Crop images based on overlapped angles: Fig. \ref{fig: crop} illustrates the process of cropping each pair of camera images based on the overlapping FoV. Redundancy occurs in the bottom two cropped images.

(c) Calculate cosine similarity: for the cropped images in the six pairs, we calculate the cosine similarity of each sample and exclude the ones without redundant instances. This provides us with a preliminary understanding of redundancy. 

(d) Create different levels of redundant training datasets to investigate how redundancy affects the inference performance: for each pair of overlapping detections, we compute a \textbf{Bounding Box Completeness Score}, which indicates how completely the bounding box (BBox) presents the instance.

Let $\mathrm{BBox}_{\mathrm{full}}$ be the original (uncropped) 2D bounding box area in the image, and let $\mathrm{BBox}_{\mathrm{clipped}}$ be the visible portion after clipping to the image boundaries.  We define:
\begin{equation}
  \mathrm{BCS}(b)
  \;=\;
  \frac{
    \mathrm{Area}\bigl(\mathrm{BBox}_{\mathrm{clipped}}(b)\bigr)
  }{
    \mathrm{Area}\bigl(\mathrm{BBox}_{\mathrm{full}}(b)\bigr)
  }\,,
\end{equation}
where $b$ indexes a candidate box.  Within each redundant group, if
\[
  \max_b \mathrm{BCS}(b)\;-\;\min_b \mathrm{BCS}(b)
  \;>\;
  \tau_{\mathrm{BCS}},
\]
We retain only the box with the higher BCS and discard the lower one; otherwise, we preserve both boxes. As the threshold $\tau_{\mathrm{BCS}}$ increases, fewer boxes are removed, thus retaining more redundancy while still preferring more complete annotations.

(e) Train YOLOv8 on each pair of overlapping camera images: from the previous step, we obtain training sets with different levels of redundancy. Next, we train the model using these training datasets and evaluate how removing redundancy affects the inference performance.

\subsection{Evaluating Redundancy in Multisource Data in Argoverse 2}

For Argoverse 2, the multisource redundancy experiment follows the same overall pipeline as the nuScenes multisource study, but is adapted to AV2’s log-based organization, surround-view sensor layout, and 3D-first annotation format. Table~\ref{tab:av2_nuscenes_method_compare} summarizes the methodological alignment between the nuScenes and AV2 multisource experiments, highlighting the shared instance-level BCS-guided pruning framework while clarifying dataset-specific differences in sensor layout, annotation space, and calibration-driven overlap pairing. 

Conceptually, the two datasets share the same core design: object-centric supervision with track-level identifiers, a BCS to compare alternative views, and a consistent YOLOv8 configuration so that observed performance changes are attributable to redundancy removal rather than model hyperparameters. In nuScenes-mini, redundancy candidates may be pre-screened using overlap-region similarity cues before BCS-guided pruning. In AV2, redundancy is defined from 3D ego-frame cuboids that are projected into each camera view to obtain 2D boxes and their BCS scores. In both cases, pruning is applied at the instance level: the lower-BCS observation is removed when the BCS gap between two overlapping views exceeds the pruning threshold $\tau_{\mathrm{BCS}}$.

On the AV2 side, multiple logs are converted into a unified multi-camera YOLO dataset, with a shared train/validation split defined over $(\textit{log\_id}, \textit{timestamp})$ keys and a global class map across categories. Camera overlap pairs are not hard-coded; instead, they are derived from calibration by estimating each camera’s yaw center and horizontal FoV from extrinsics and intrinsics, then computing angular overlap on the viewing circle to select overlapping camera pairs for pruning. Using this overlap graph, the experiment constructs a baseline (unpruned) dataset and a family of BCS-pruned datasets that differ only in the pruning threshold $\tau_{\mathrm{BCS}}$. Among these, $\tau_{\mathrm{BCS}}=0.5$ provides a balanced operating point in our sweep, removing redundant labels while keeping performance close to the unpruned baseline.

\begin{table*}[t]
\centering
\caption{Comparison between the redundancy experiments on nuScenes and AV2 datasets. Both apply instance-level, BCS-guided pruning; AV2 adds calibration-driven overlap pairing and 3D-to-2D projection. All the experiments were conducted on a PC equipped with one NVIDIA GeForce 4090 RTX GPU and 24 GB of memory. Model configurations are fixed across experiments.}
\setlength{\tabcolsep}{3pt}
\renewcommand{\arraystretch}{1.08}
\footnotesize
\begin{tabularx}{\textwidth}{p{2.7cm} X X}
\toprule
\textbf{Aspect} & \textbf{nuScenes multisource} & \textbf{AV2 multisource} \\
\midrule
Sensor layout &
6 calibrated surround-view cameras with predefined overlapping FoVs. &
9 calibrated cameras (7 ring + 2 front stereo); surround-view configuration with denser overlap. \\
\midrule
Source of supervision &
Object identity inferred implicitly from dataset annotations and temporal alignment. &
Explicit object identity via \texttt{track\_uuid} from 3D ego-frame cuboids. \\
\midrule
Annotation space &
2D image-space object-centric observations derived from overlap regions. &
3D-first cuboids in a unified ego frame, projected into each camera view to obtain 2D observations. \\
\midrule
Redundancy unit &
Multiple 2D observations of the same object instance across overlapping cameras. &
Multiple projected 2D observations originating from the same 3D object instance across cameras. \\
\midrule
Overlap pairing &
Fixed camera overlap pairs defined by dataset specification. &
Camera overlap pairs computed from calibration (yaw center and horizontal FoV) using angular overlap. \\
\midrule
BCS computation &
Boundary-aware 2D coverage score computed on valid image regions. &
Boundary-aware 2D coverage score computed after projecting 3D cuboids and clipping to image bounds. \\
\midrule
Pruning rule &
Drop the lower-BCS observation if the BCS gap exceeds $\tau_{\mathrm{BCS}}$. &
Same instance-level rule applied to projected views. \\
\midrule
Training protocol &
YOLOv8 with matched hyperparameters to isolate redundancy effects. &
Same YOLOv8 configuration; baseline (unpruned) and BCS-pruned datasets differ only by $\tau_{\mathrm{BCS}}$. \\
\bottomrule
\end{tabularx}

\label{tab:av2_nuscenes_method_compare}
\end{table*}

\subsection{Evaluating Redundancy in Multimodal Data}

Fig. \ref{fig: cross-modal redundancy} shows an example of multimodal data redundancy in the nuScenes dataset. The LiDAR on top and the front camera capture the same objects. This experiment investigates the redundancy between these two sensors:

\begin{figure}
  \centering
  \includegraphics[width=0.9\linewidth]{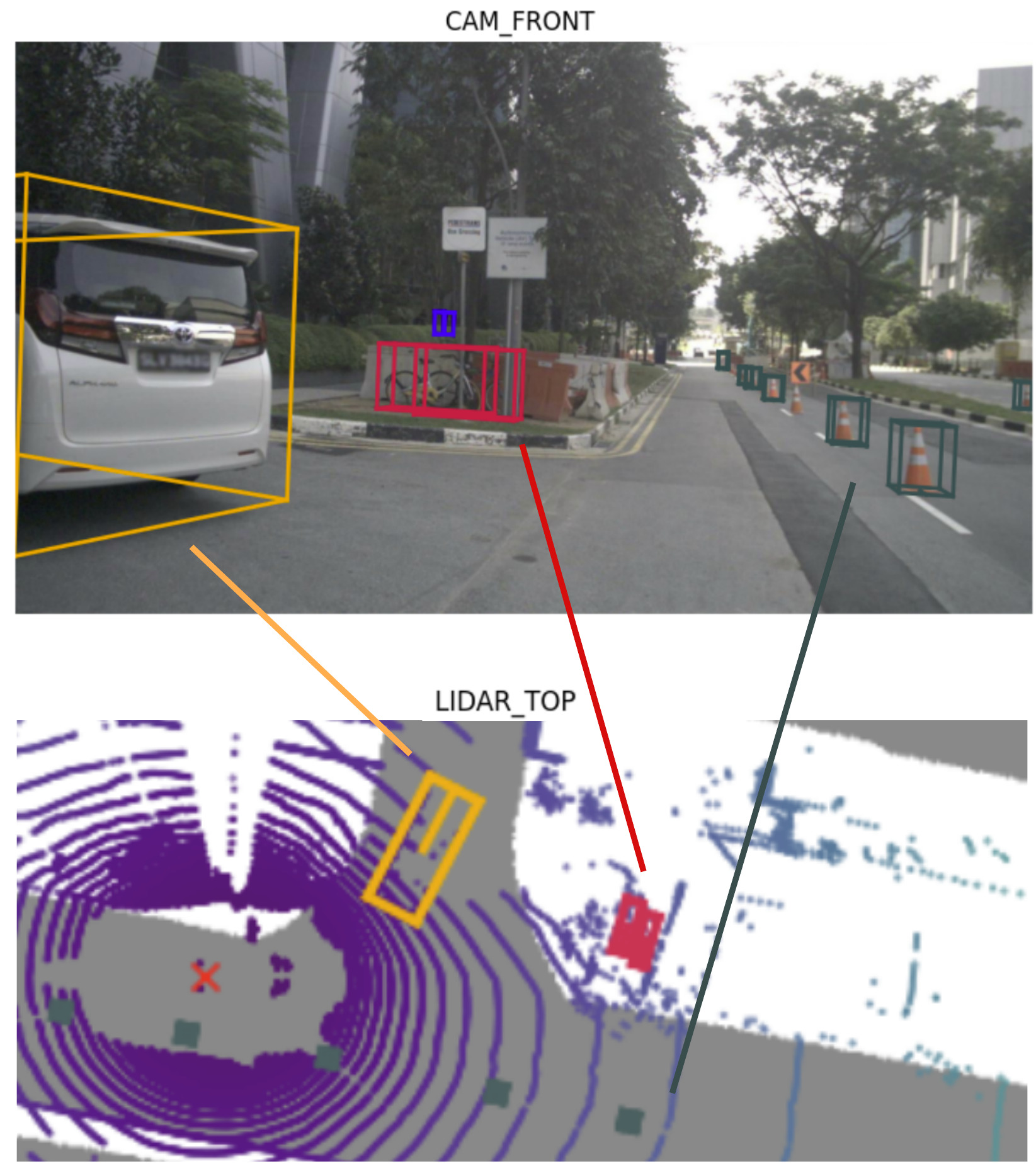}
  \caption{Example of cross-modal redundancy between the front camera (upper) and the top LiDAR (lower) in the nuScenes dataset. Colored lines indicate the corresponding objects observed by both sensors.}
  \label{fig: cross-modal redundancy}
\end{figure}

(a) Run the full camera-LiDAR fusion detection model \cite{TimKie} to get baseline 3D bounding boxes. Check the overlapping detections between this and the detection using only LiDAR data as redundancy. Let $\mathcal{B}_{\mathrm{LiDAR}}$ be the set of boxes detected using LiDAR only, and $\mathcal{B}_{\mathrm{base}}$ the set detected by the image-LiDAR fusion. The \emph{redundancy ratio} for one frame is
\begin{equation}
  \mathrm{RR}
  \;=\;
  \frac{\bigl|\{\,b\in\mathcal{B}_{\mathrm{base}}\mid\exists\,b'\in\mathcal{B}_{\mathrm{LiDAR}}:\mathrm{IoU}(b,b')\ge\theta\}\bigr|}
       {|\mathcal{B}_{\mathrm{base}}|}\,.
\end{equation}

(b) Compute each box’s 3D centroid in the LiDAR frame, and measure its distance from the ego sensor: 

\[
  \mathbf{c}(b) \;=\; \frac{1}{8}\sum_{i=1}^{8} \mathbf{v}_i
  \quad\text{and}\quad
  d(b) \;=\; \bigl\lVert \mathbf{c}(b)\bigr\rVert_2,
\]
where \(\{\mathbf{v}_i\}\) are the eight corner vertices of \(b\). We then set a single pruning threshold $T_{\mathrm{dist}}$ (meters). All boxes whose centroids lie within this distance are removed:

\begin{equation}
  \mathcal{B}_{\mathrm{pruned}}
  \;=\;
  \bigl\{\,\mathcal{B}_{\mathrm{LiDAR}} \mid d(\mathcal{B}_{\mathrm{LiDAR}}) \;\ge\; T_{\mathrm{dist}}\bigr\}.
\end{equation}

By sweeping \(T_{\mathrm{dist}}\), we could trace how many boxes are pruned and what fraction of true detections is lost, thus choosing a trade-off point that maximally reduces redundancy without excessively harming detection performance. This distance threshold is chosen based on statistical results explained in the next section.

(c) Run LiDAR detection again using the partially removed LiDAR data and observe how removing redundancy affects the performance. Let $\mathcal B_{\mathrm{pruned}}$ be the set after pruning. We define the \emph{lost ratio} $\ell$ as the fraction of baseline boxes removed due to removing the redundancy:
\[
  \ell
  \;=\;
  \frac{\bigl|\mathcal B_{\mathrm{base}}\setminus \mathcal B_{\mathrm{pruned}}\bigr|}
       {\bigl|\mathcal B_{\mathrm{base}}\bigr|}
  \;=\;
  1 \;-\;
  \frac{\bigl|\mathcal B_{\mathrm{base}}\cap \mathcal B_{\mathrm{pruned}}\bigr|}
       {\bigl|\mathcal B_{\mathrm{base}}\bigr|}.
\]

\section{Results}
\label{sec: result}

\subsection{Overall Results}

We conduct the multisource redundancy experiments on two datasets under a unified protocol. We first establish the workflow on nuScenes and then repeat the same redundancy pruning procedure on Argoverse~2 under a controlled evaluation setup. Table \ref{tab:av2_overview_baseline_tau05} summarizes the 2D detection performance. For each dataset, we report results for the unpruned baseline (trained and evaluated using all valid camera views) and the results using pruned training datasets.

\begin{itemize}
    \item \textbf{nuScenes.} The baseline detector was evaluated on $1,401$ images containing $12,286$ labeled object instances across $14$ categories. Overall, the detector achieved a box precision of $0.780$, a recall of $0.630$, and an mAP$50$ of $0.720$. These results serve as the nuScenes baseline for subsequent multisource redundancy analysis. Using a threshold of $\tau_{\mathrm{BCS}}=0.8$ on pair 4, which removes approximately $7\%$ of the baseline, detection performances improved and became the best among all pairs. 
    \item \textbf{Argoverse~2.} For AV2, the baseline YOLOv8 was evaluated on the full training set, which contains $187,265$ projected 2D labels corresponding to $95,266$ 3D object tracks across $19$ classes. Using a threshold of $\tau_{\mathrm{BCS}}=0.5$, which removes $9,442$ labels (about 5.0\% of the baseline), precision performance remained close, with only modest reductions in recall and mAP${50}$.
\end{itemize}

Compared to the AV2 baseline, pruning at $\tau_{\mathrm{BCS}}=0.5$ yields a modest reduction in recall and mAP, while precision remains slightly higher than the unpruned baseline.
Overall, detection performance remains close to the baseline despite removing redundant camera-level supervision.
A more detailed analysis of the performance--redundancy trade-off across pruning thresholds is presented in the following sections.

\begin{table}[t]
\centering
\footnotesize
\caption{Multisource detection performance with and without BCS-guided pruning}
\begin{tabular}{p{0.9cm} p{2.1cm}  c c c c}
\toprule
\textbf{Dataset} & \textbf{Experiment}  &
\textbf{Prec.} & \textbf{Rec.} &
\textbf{mAP$50$} &
\textbf{mAP$50:95$} \\
\midrule

nuScenes & Train unpruned, Eval unpruned  & 0.780 & 0.630 & 0.720 & 0.450 \\

nuScenes & Train pruned ($\tau_{\mathrm{BCS}}{=}0.8$ on pair 4), Eval unpruned  & 0.867 & 0.800 & 0.770 & 0.638 \\

AV2 & Train unpruned, Eval unpruned  & 0.815 & 0.565 & 0.640 & 0.447 \\

AV2 & Train pruned ($\tau_{\mathrm{BCS}}{=}0.5$), Eval unpruned  & 0.818 & 0.557 & 0.622 & 0.433 \\

\bottomrule
\end{tabular}

\label{tab:av2_overview_baseline_tau05}
\end{table}

\subsection{Impact of Redundancy in Multisource Data in nuScenes}

\begin{figure}
    \centering
    \includegraphics[width=1.0\linewidth]{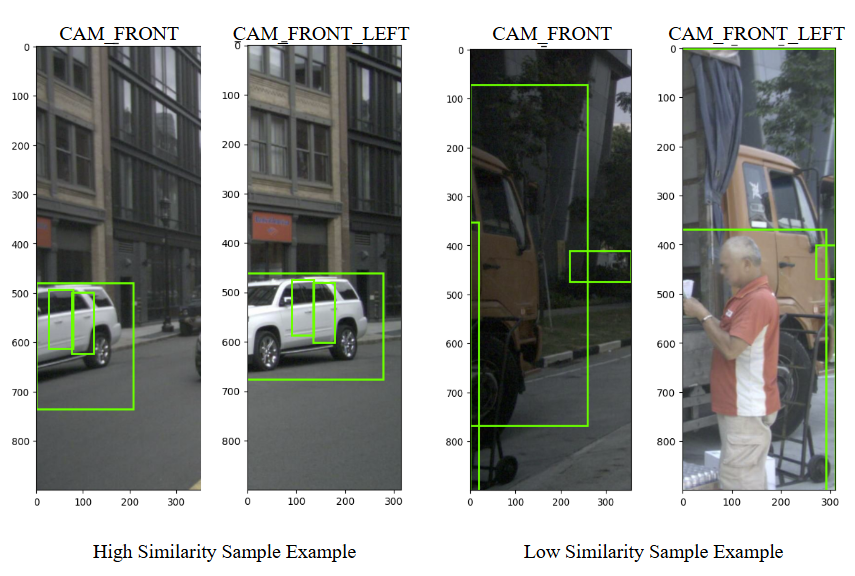}
    \caption{Examples of high and low similarity samples in nuScenes.}
    \label{fig: sim}
\end{figure}

For multisource redundancy evaluation, Fig. \ref{fig: sim} shows an example of high-similarity (left) and low-similarity (right) views in the front and front left cameras in nuScenes. In high-similarity examples (left), the detected objects and background almost provide the same information. By contrast, the low-similarity ones (right) present more complementary detections. 

\begin{figure*}
    \centering
    \includegraphics[width=0.8\linewidth]{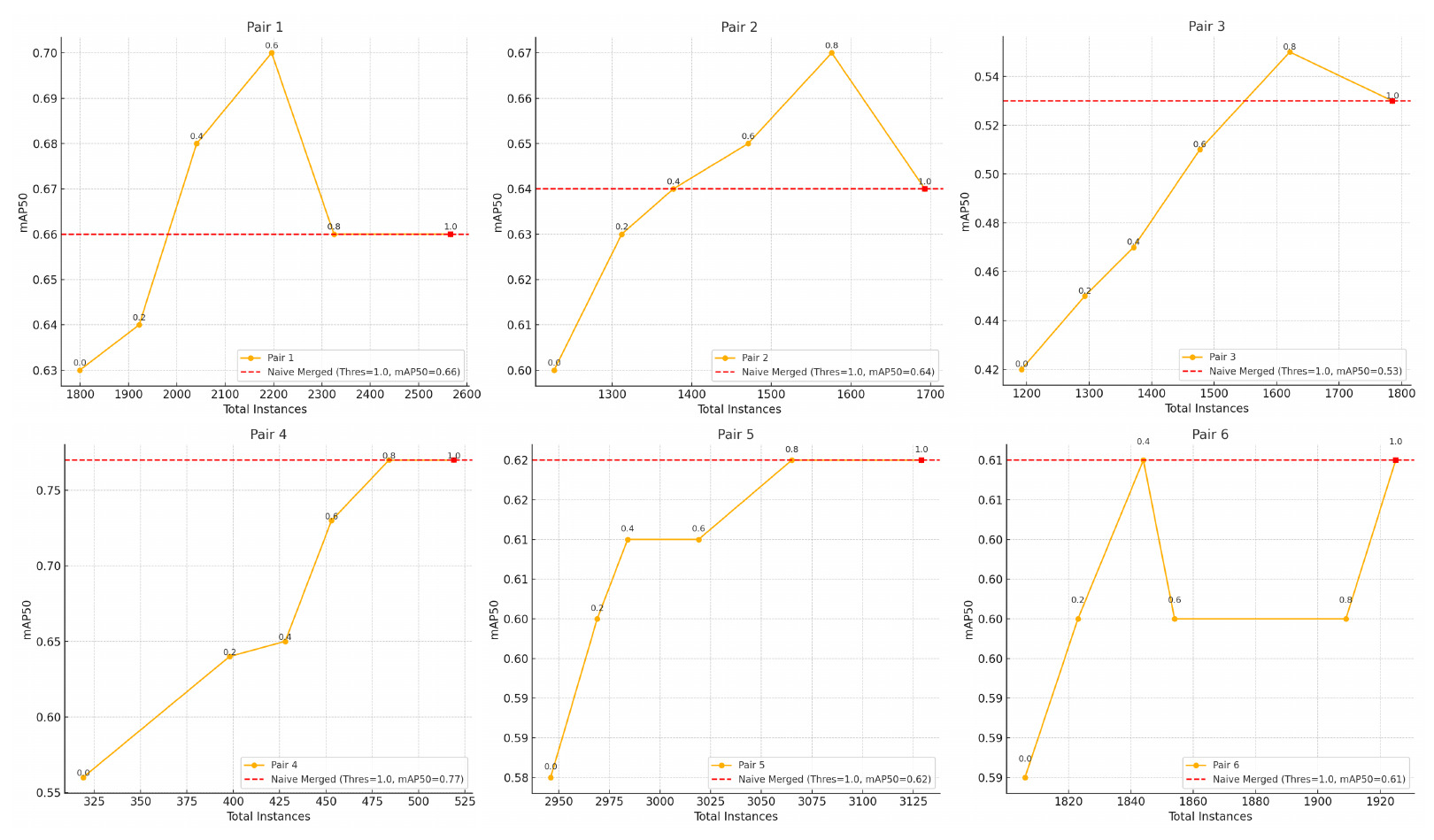}
    \caption{Effects of removing redundant instances in overlapping FoV images on mAP$50$ in nuScenes.}
    \label{fig: full pair}
\end{figure*}

Fig. \ref{fig: full pair} shows how removing redundancy affects detection performance on six pairs in Fig. \ref{fig: camera}. For each pair, the thresholds are set from 0.0 to 1.0, with a 0.2 interval. As the threshold increases, more instances in the training dataset are retained, meaning that more redundancy is preserved until the 1.0 threshold retains all instances for training. The trends reveal that a less redundant training dataset can achieve or even surpass the performance level of using the full training dataset.  

\subsection{Impact of Redundancy in Multisource data in Argoverse 2}
\label{sec:av2_redundancy}
To examine whether multisource redundancy generalizes beyond nuScenes, we repeat the instance-level BCS-guided pruning experiment on AV2. With different $\tau_{\mathrm{BCS}}$ thresholds, Table~\ref{tab:av2_tau_sweep} summarizes the redundancy--performance trade-off on AV2. 

\begin{table}[t]
\centering
\caption{AV2 multisource BCS pruning sweep. Training uses different $\tau_{\mathrm{BCS}}$ values while evaluation is fixed to the same unpruned validation split. ``Del.'' and ``Rem.'' denote deleted and remaining camera-level 2D labels in the training set. ``Tracks'' is the number of unique 3D object tracks observed during training.}
\setlength{\tabcolsep}{2.6pt} 
\renewcommand{\arraystretch}{1.04}
\footnotesize
\begin{tabular}{l c r r r c c c}
\toprule
\textbf{Train} & $\tau_{\mathrm{BCS}}$ &
\textbf{Del.} & \textbf{Rem.} & \textbf{Tracks} &
\textbf{Prec.} & \textbf{Rec.} & \textbf{mAP${50}$} \\
\midrule
Baseline & --  & 0       & 187{,}265 & 95{,}266 & 0.815 & 0.565 & 0.640 \\
Pruned   & 0.1 & 16{,}056 & 171{,}209 & 95{,}266 & 0.811 & 0.531 & 0.604 \\
Pruned   & 0.2 & 14{,}345 & 172{,}920 & 95{,}266 & 0.788 & 0.544 & 0.605 \\
Pruned   & 0.3 & 12{,}810 & 174{,}455 & 95{,}266 & 0.826 & 0.538 & 0.617 \\
Pruned   & 0.4 & 11{,}018 & 176{,}247 & 95{,}266 & 0.823 & 0.545 & 0.618 \\
Pruned   & 0.5 & 9{,}442  & 177{,}823 & 95{,}266 & 0.818 & 0.557 & 0.622 \\
Pruned   & 0.6 & 7{,}651  & 179{,}614 & 95{,}266 & 0.812 & 0.554 & 0.630 \\
\bottomrule
\end{tabular}

\label{tab:av2_tau_sweep}
\end{table}

Across the sweep, BCS-based pruning removes a non-trivial fraction of camera-level labels (approximately 4.1--8.6\%, depending on $\tau_{\mathrm{BCS}}$), while the number of unique 3D object tracks observed during training remains constant (95{,}266 tracks). This confirms that redundancy is primarily introduced by overlapping redundant multi-camera observations rather than by removing distinct object instances.

Detection performance remains within a narrow margin of the unpruned baseline across a broad range of thresholds. In particular, $\tau_{\mathrm{BCS}}=0.5$ discards 9{,}442 labels (about 5.0\%) while maintaining comparable precision (0.818 vs.\ 0.815), with only modest reductions in recall and mAP${50}$. Overall, these trends mirror the nuScenes results, indicating that redundant multisource supervision can be reduced substantially with minimal accuracy cost under a controlled, fixed evaluation protocol.

Compared to nuScenes, AV2 introduces additional implementation complexity due to its log-based organization, calibration-driven overlap estimation, and 3D-first annotation pipeline. Despite these differences, the consistent performance--redundancy behavior suggests that multisource redundancy is an intrinsic property of multi-camera supervision rather than a dataset-specific artifact.

\subsection{Impact of Redundancy in Multimodal Data}

\begin{figure}
    \centering
    \includegraphics[width=1.0\linewidth]{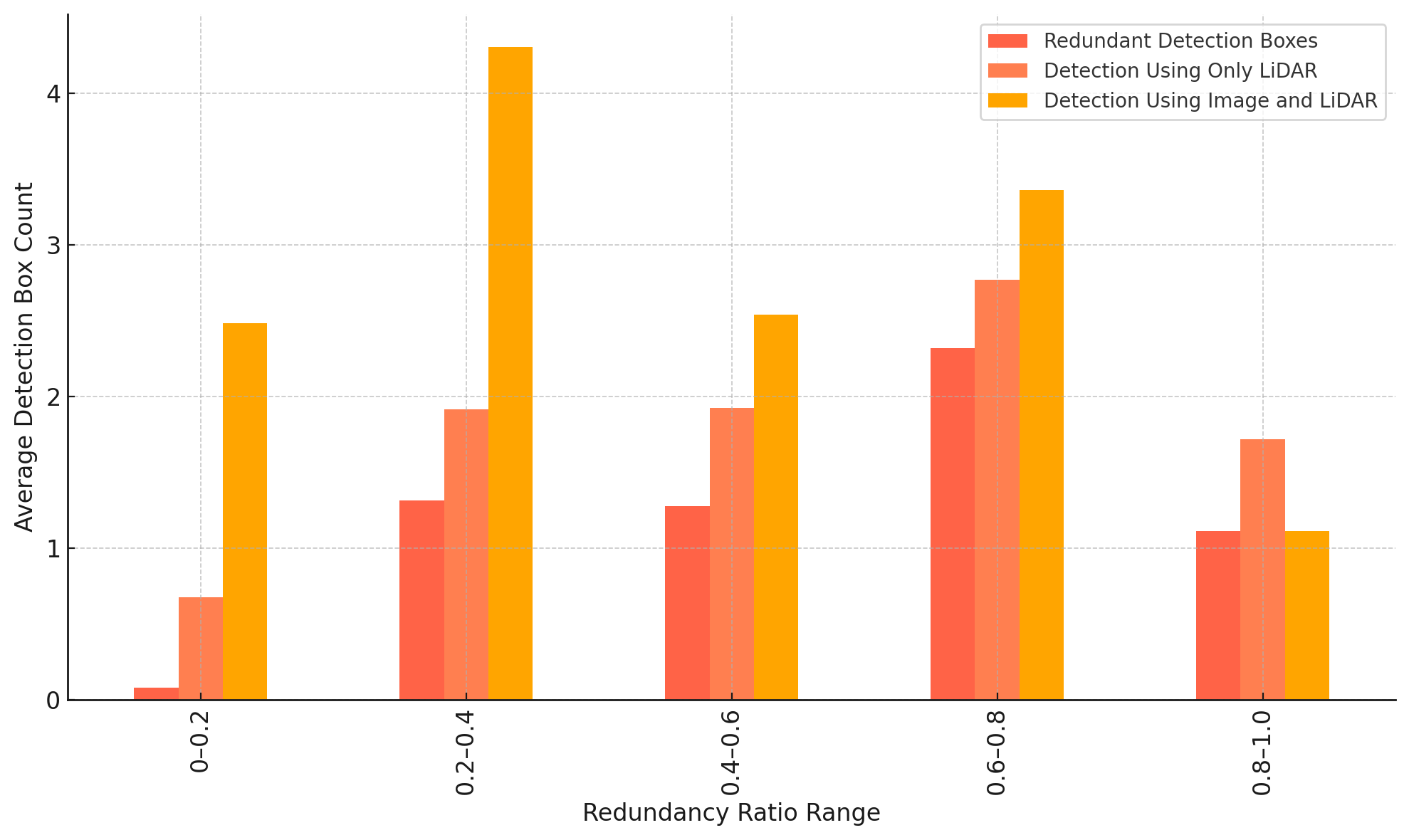}
    \caption{Redundancy in LiDAR and image data in nuScenes.}
    \label{fig: lidar-red}
\end{figure}

\begin{figure}
    \centering
    \includegraphics[width=0.8\linewidth]{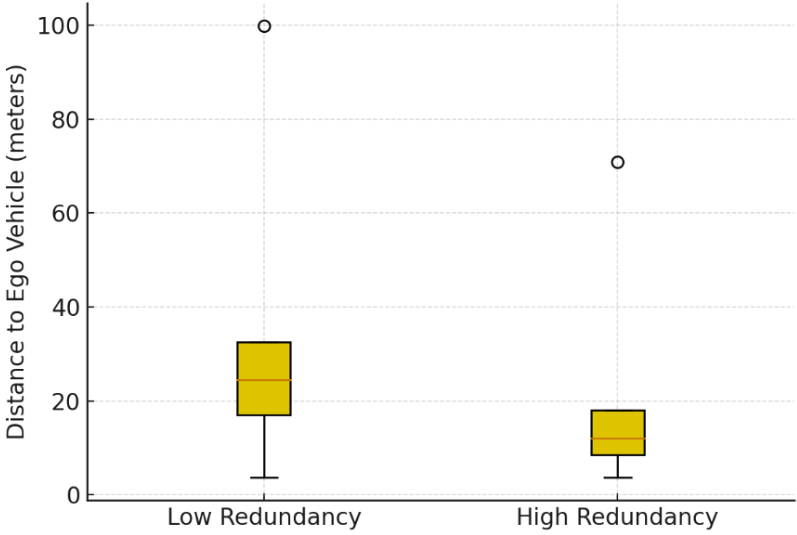}
    \caption{T-test of distance threshold and cross-modal redundancy in nuScenes.}
    \label{fig: distance stat}
\end{figure}

\begin{figure}
    \centering
    \includegraphics[width=0.9\linewidth]{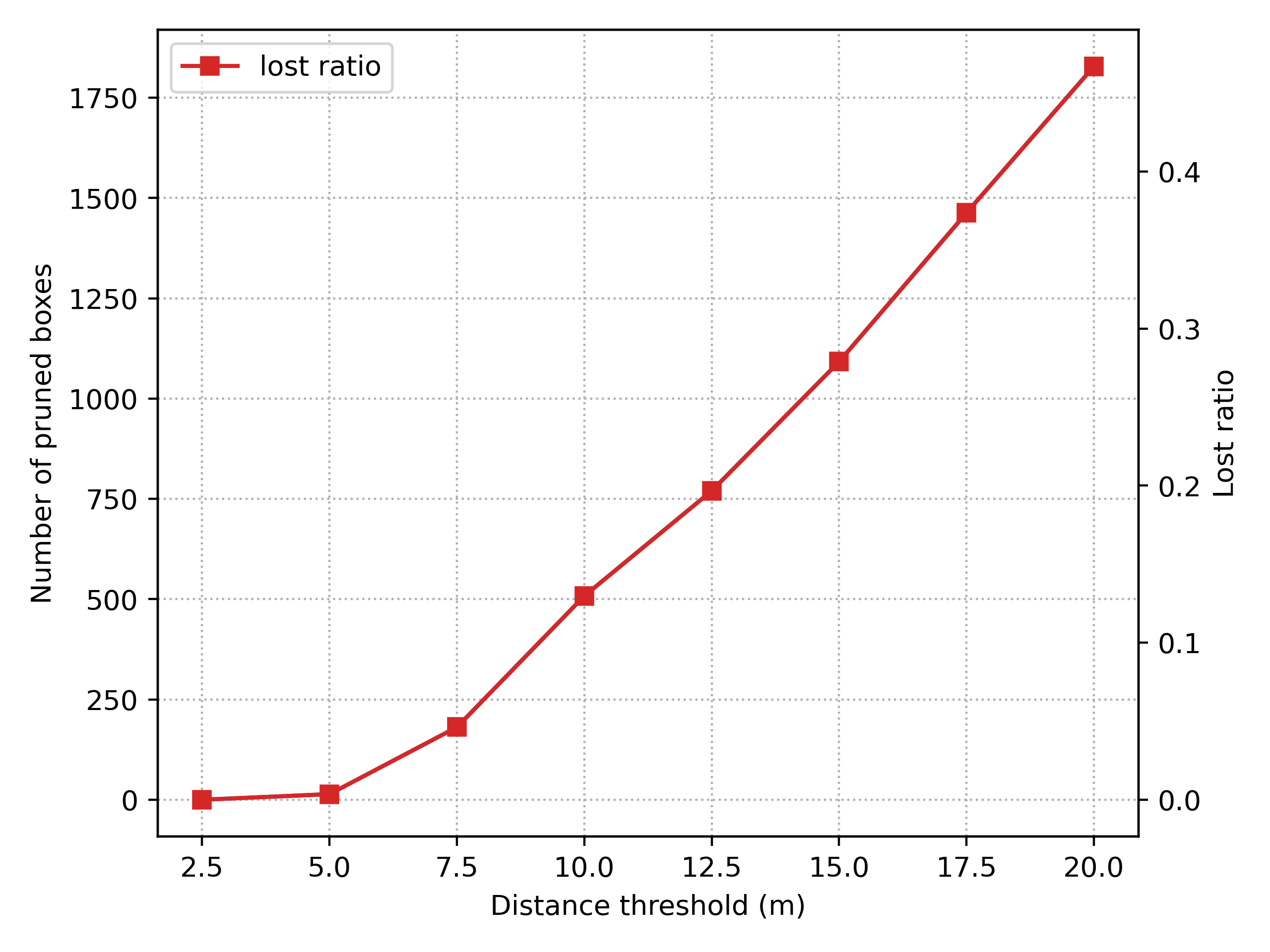}
    \caption{The effects of removing close-range redundant LiDAR data in nuScenes on detection performance.}
    \label{fig: distance_pruning_curve}
\end{figure}

For multimodal redundancy evaluation, Fig. \ref{fig: lidar-red} shows the distribution of redundancy across the Lidar and image detection in nuScenes. We further group and compare the high-redundancy and low-redundancy samples. T-test results of distance threshold \(T_{\mathrm{dist}}\) and cross-modal redundancy (p-value=1.17e-76) in Fig. \ref{fig: distance stat} suggest that the objects with high cross-modal redundancy remain very close to the ego-vehicle. This supports the removal of close-range LiDAR data as redundancy. Fig. \ref{fig: distance_pruning_curve} shows that the detections can be almost unaffected by choosing a reasonable threshold and removing near-range LiDAR points. In conclusion, for multimodal data in object detection, closer-range LiDAR data is more redundant. While removing them could have little impact, the efficiency can be improved due to the decrease in data points to be processed. 

\section{Discussion}
\label{sec: discussion}

We demonstrate that redundancy exists in (1) multisource and single modality data, which is the multi-view camera images; and (2) multisource and multimodal data, which is the image and LiDAR data, using qualitative illustrations and quantitative statistics in three experiments.

\subsection{Redundancy in Multisource Data}

This existing multisource redundancy in two datasets is not surprising, as the autonomous driving system requires a certain level of redundancy to cross validate and ensure safety \cite{liu2021Multi-sensorRedundancy, qian20223d, he2020sensorRedun, zhang2022sensor, song2019apollocar3d}. From a data quality perspective, however, simply accumulating more overlapping views of the same physical object does not necessarily provide more informative supervision. Instead, it can introduce low-informative redundancy that increases dataset size and computational cost without improving model behavior. Accordingly, our multisource experiments focus on (1) modeling and measuring redundancy at the instance level for object detection, and (2) removing low-informative yet computation-consuming redundancy while preserving complementary observations that contribute to performance.

From the nuScenes multisource experiment (Fig. \ref{fig: full pair}), introducing the Bounding Box Completeness Score effectively reduces redundancy by preferentially removing lower-quality redundant boxes while preserving complementary views when overlapping observations are of similar quality. In several camera pairs, models trained on a less redundant subset achieve comparable, and in some cases slightly higher, detection performance than models trained on the full unpruned dataset. This observation highlights that redundant multisource supervision may provide diminishing returns once at least one high-quality view of an object is available, reinforcing the importance of data quality over sheer quantity.

The AV2 results further support this conclusion. Across a sweep of pruning thresholds $\tau_{\mathrm{BCS}} \in [0.1, 0.6]$ (Table~\ref{tab:av2_tau_sweep}), the pruning procedure removes between 7,651 and 16,056 camera-level labels out of 187,265 candidates (approximately 4.1\%--8.6\%), while the number of unique 3D object tracks observed during training remains constant at 95,266. This indicates that multisource redundancy in AV2 primarily arises from overlapping camera projections of the same physical objects, rather than from the presence of distinct instances.

Despite this substantial reduction in supervision, detection performance remains close to the unpruned baseline across the entire sweep. For example, at $\tau_{\mathrm{BCS}}=0.5$, the model discards 9,442 labels (about 5.0\%) while maintaining comparable precision (0.818 vs.\ 0.815) and only modest reductions in recall and mAP${50}$. These trends closely mirror the nuScenes results and demonstrate that instance-level redundancy pruning yields a more compact training signal without substantially degrading downstream detection performance.

Finally, BCS-guided pruning preserves the relative representation of instances across overlapping camera pairs. The proportion of instances contributed by key camera overlaps remains similar before and after pruning, suggesting that redundancy is reduced without introducing strong viewpoint imbalance. Overall, these findings suggest that multisource redundancy should be treated as a controllable data quality dimension: by explicitly managing redundant supervision at the instance level, it is possible to improve dataset efficiency while maintaining robust perception performance.

\subsection{Redundancy in Multimodal Data}

The evaluation of image-LiDAR redundancy reveals that high redundancy ratios tend to occur for objects located close to the ego-vehicle, where dense LiDAR returns. This suggests that these instances are being detected equally well by both modalities and therefore contribute little new signal when both are included. This finding calls back to the multisource redundancy that selectively removing redundancy could maintain the performances, while offering other benefits, such as reducing the training dataset size.

The BCS in the first experiment also guides us in looking into the relationship between the distance to the ego vehicle and the 3D bounding box size. Zhang et al. find that the distance captured in the point cloud is a crucial factor for achieving a high point cloud quality \cite{zhang2024igo}. Large, well-structured objects are easy for sensors to detect. Ideally, the further the object is located, the smaller the 3D bounding box size is. However, the bounding box quality remains a concern. From qualitative checks, the boxes could not accurately present the object size, despite correctly providing the centroid of the location. This was also observed in Kim et al.'s study on label quality \cite{kim2025automatic}. They pointed out that many boxes are labeled as either too small or too big, resulting in low-quality datasets.

\section{Conclusion and Future Work}

\label{sec:con}
In this work, we first put forward a research gap: while AVs collect increasingly large volumes of $M^2$ data, redundancy as a crucial DQ dimension has been overlooked. To address this, we model and measure redundancy across $M^2$ data using nuScenes and Argoverse 2 datasets in object detection tasks using YOLOv8. 

The results prove that selectively removing redundant multisource image data leads to measurable improvements in detection performance. In nuScenes, removing redundant labels preserves accuracy and frequently improves mAP${50}$, with representative gains including Pair~1 increasing from $0.66$ to $0.70$, Pair~2 from $0.64$ to $0.67$, and Pair~3 from $0.53$ to $0.55$, while other pairs (4, 5, 6) match their respective baselines even under more aggressive pruning. In AV2, approximately $4.1$--$8.6\%$ labels are removed, and detection performance remains near the baseline $0.64$ mAP${50}$, with the best redundancy--accuracy trade-off achieved at $\tau_{\mathrm{BCS}}{=}0.6$. 

Additionally, our analysis of multimodal data in image and LiDAR point clouds highlights existing redundancy issues that traditional pipelines do not consider, revealing opportunities to make perception systems more efficient without compromising reliability. Finally, further exploration of redundancy in AVs will focus on the following perspectives: 


\begin{enumerate}
    \item This paper has initially explored data redundancy in AD using the nuScenes and AV2 datasets, demonstrating that partially removing redundant data could improve model performance. We will extend this evaluation across diverse, unlabeled, and larger-scale datasets, suitable and SOTA models.

    \item We aim to thoroughly evaluate the impacts of redundancy under varied conditions. For example, we will investigate how redundancy patterns shift across driving environments (e.g., dense urban vs. highway) and other factors such as lighting, weather, and sensor visibility.
    
    \item Additional modalities, such as RADAR or user-generated data, introduce complexity, potentially revealing more redundancy. Integrating these modalities could further refine how to model and measure redundancy, as well as other important data quality dimensions. 
    
    \item Different AV tasks prioritize data quality differently. Other tasks, such as prediction and planning, may emphasize different metrics over redundancy, which is critical for safe decision-making in end-to-end autonomous driving systems.

\end{enumerate}

\section*{Acknowledgment}

The authors would like to thank Dr. Junhua Ding from the Department of Data Science, Drs. Heng Fan and Yunhe Feng from the Department of Computer Science and Engineering at the University of North Texas for their precious feedback and suggestions. This work was supported in part by the National Science Foundation (NSF) under Award 2505686.

\bibliographystyle{IEEEtran}
\bibliography{reference}

\end{document}